\documentclass[runningheads]{llncs}
\usepackage[T1]{fontenc}
\usepackage{graphicx,verbatim}
\usepackage{amsmath}
\usepackage{amssymb}
\usepackage{booktabs}
\usepackage{array}
\usepackage{url}
\usepackage{tikz}
\usepackage{pgfplots}
\usepackage[table]{xcolor}
\usepackage{multirow}
\usepackage{listings}
\usepackage{microtype}
\usetikzlibrary{arrows.meta,positioning,fit,calc}
\pgfplotsset{compat=1.18}
\usepgfplotslibrary{groupplots}
\definecolor{jsonkey}{rgb}{0.0,0.1,0.6}
\definecolor{jsonstr}{rgb}{0.6,0.1,0.1}
\definecolor{jsonkw}{rgb}{0.5,0.0,0.5}
\newcommand{\std}[1]{\,{\scriptstyle\pm\,#1}}

\begin{document}

\title{VIVID-Med: LLM-Supervised Structured Pretraining for Deployable Medical ViTs}
\titlerunning{VIVID-Med: LLM-Supervised Pretraining for Medical ViTs}

\author{Xiyao Wang\inst{1} \and
Xiaoyu Tan\inst{2} \and
Yang Dai\inst{1} \and
Yuxuan Fu\inst{1} \and
Shuo Li\inst{3} \and
Xihe Qiu\inst{1}\thanks{Corresponding author.}}
\authorrunning{Wang et al.}
\institute{Shanghai University of Engineering Science, China \and
National University of Singapore, Singapore \and
Case Western Reserve University, United States}

\maketitle

\begin{abstract}
Vision-language pretraining has driven significant progress in medical image analysis. However, current methods typically supervise visual encoders using one-hot labels or free-form text, neither of which effectively captures the complex semantic relationships among clinical findings. In this study, we introduce \textbf{VIVID-Med}, a novel framework that leverages a frozen large language model (LLM) as a structured semantic teacher to pretrain medical vision transformers (ViTs). \textbf{VIVID-Med} translates clinical findings into verifiable JSON field-state pairs via a Unified Medical Schema (UMS), utilizing answerability-aware masking to focus optimization. It then employs Structured Prediction Decomposition (SPD) to partition cross-attention into orthogonality-regularized query groups, extracting complementary visual aspects. Crucially, the LLM is discarded post-training, yielding a lightweight, deployable ViT-only backbone. We evaluated \textbf{VIVID-Med} across multiple settings: on CheXpert linear probing, it achieves a macro-AUC of 0.8588, outperforming BiomedCLIP by {+}6.65 points while using 500$\times$ less data. It also demonstrates robust zero-shot cross-domain transfer to NIH ChestX-ray14 (0.7225 macro-AUC) and strong cross-modality generalization to CT, achieving 0.8413 AUC on LIDC-IDRI lung nodule classification and 0.9969 macro-AUC on OrganAMNIST 11-organ classification. \textbf{VIVID-Med} offers a highly efficient, scalable alternative to deploying resource-heavy vision-language models in clinical settings.

\end{abstract}

\section{Introduction}

Learning transferable visual representations is central to robust medical image analysis. Existing approaches, such as supervised pretraining~\cite{dosovitskiy2020image}, self-supervised methods (MAE~\cite{he2022masked}, DINO~\cite{caron2021emerging,simeoni2025dinov3}), general vision-language models~\cite{radford2021learning,jia2021scaling,li2022blip,alayrac2022flamingo,driess2023palm,liu2023visual}, and medical vision-language pretraining (ConVIRT~\cite{zhang2022contrastive}, GLoRIA~\cite{huang2021gloria}, MedCLIP~\cite{wang2022medclip}, BiomedCLIP~\cite{zhang2023biomedclip}, CheXzero~\cite{tiu2022expert}), have driven strong progress. Recent work has also explored multimodal data integration~\cite{qiu2026mimar} and few-shot medical image translation~\cite{qiu2026sru} to enhance representation learning. However, existing methods predominantly supervise the visual encoder using one-hot labels or free-form text. Consequently, these learned representations often fail to explicitly encode the complex semantic relationships among clinical findings. For example, conditions like \emph{pleural effusion} and \emph{pulmonary edema} frequently co-occur, share radiographic signs, and are pathophysiologically linked. A one-hot vector treats them as strictly orthogonal, while free-text descriptions mention them in highly variable phrasing, masking their underlying clinical relatedness.

Results from recent advancements in large language models (LLMs) suggest an elegant solution to this semantic gap. A pretrained LLM inherently embeds related clinical concepts in a continuous, structured space where proximity reflects actual clinical relatedness. In this study, we leverage this insight to develop a novel semantic distillation strategy for medical vision transformers (ViTs). We present the discovery that a ViT carefully optimized to align with the structured output space of a \emph{frozen} LLM can learn highly transferable representations, eliminating the need to retain the resource-heavy LLM during inference.

We propose \textbf{VIVID-Med} (\textbf{V}erifiable \textbf{I}nstruction-driven \textbf{V}isual \textbf{I}ntelligence \textbf{D}eployment for \textbf{Med}ical ViT), a framework that executes this strategy by using a \emph{frozen} LLM as a structured semantic teacher. To bridge the visual and linguistic domains effectively, \textbf{VIVID-Med} introduces two key technical components. First, the \textbf{Unified Medical Schema (UMS)} converts raw clinical findings into verifiable JSON field-state pairs. UMS utilizes answerability-aware masking to filter out non-informative gradients from unassessable findings, focusing optimization on clinically meaningful signals. Second, \textbf{Structured Prediction Decomposition (SPD)} partitions the cross-attention mechanism into orthogonality-regularized query groups via Q-Former-style~\cite{li2023blip} modules, extending knowledge distillation~\cite{hinton2015distilling} with structured decomposition. Crucially, keeping the LLM frozen provides a stable optimization target, which is particularly critical for learning representations of long-tail clinical findings.

\textbf{VIVID-Med} draws inspiration from instruction-driven methods such as ViTP \cite{li2025visual}, which also embed a ViT into a trainable vision-language model. However, ViTP relies on free-form text generation with random token dropping and requires the LLM to remain active during deployment. In contrast, \textbf{VIVID-Med} offers a more structured and deployable methodology. By utilizing verifiable JSON formats and structured query decomposition rather than random masking, and by completely discarding the LLM post-training, \textbf{VIVID-Med} produces a lightweight, deployable ViT-only backbone.

\textbf{Contributions.} In summary, our main contributions are as follows:
\begin{enumerate}
    \item We introduce a novel frozen-LLM distillation framework that produces a highly transferable and easily deployable ViT-only backbone.
    \item We propose UMS, a structured JSON supervision method featuring field query training and answerability-aware masking to focus optimization.
    \item We design SPD, a multi-group cross-attention projector with orthogonality regularization that efficiently decomposes visual features.
    \item We demonstrate the effectiveness of \textbf{VIVID-Med} through comprehensive evaluations, including CheXpert linear probing, zero-shot cross-domain transfer to NIH Chest X-ray14, cross-modality CT transfer (LIDC-IDRI, OrganAMNIST), and a detailed single-variable ablation chain.
\end{enumerate}

\section{Method}

\begin{figure}[t]
\centering
\includegraphics[width=\textwidth]{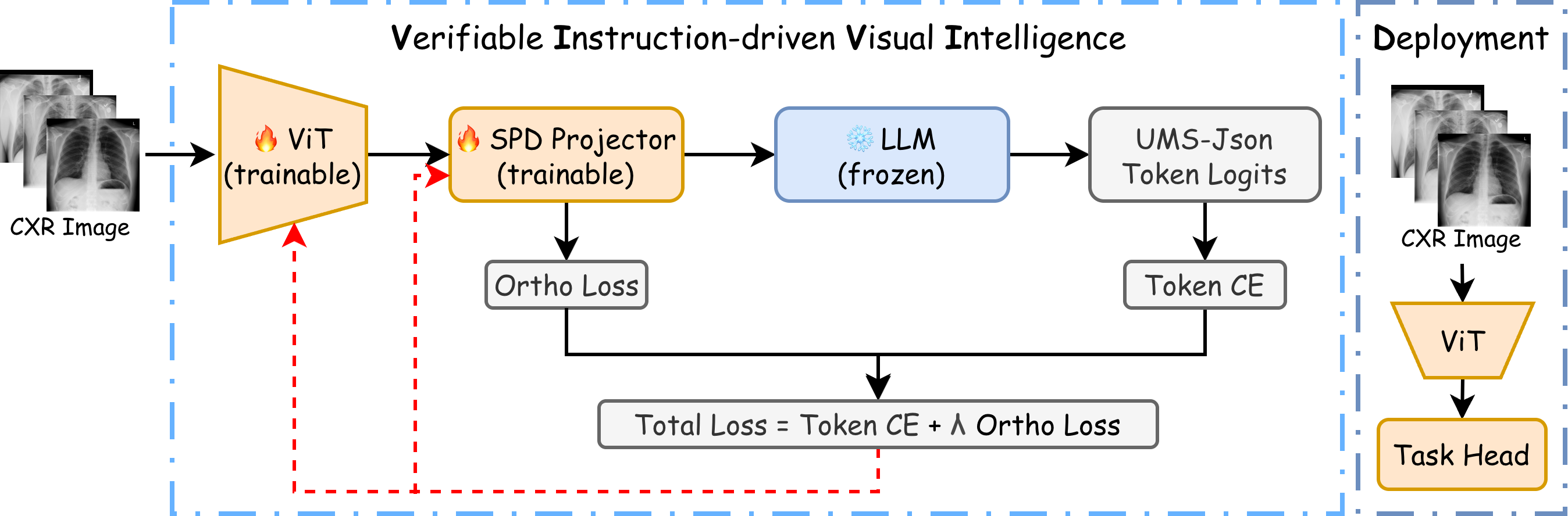}
\caption{\textbf{VIVID-Med} overview. The frozen LLM provides structured token supervision during training only. Gradients update the parameters of the ViT and SPD projector, while the deployed model retains only the lightweight ViT backbone.}
\label{fig:pipeline}
\end{figure}

Given a training set $\mathcal{D} = \{I_j, C_j\}_{j=1}^N$ of medical images $I_j$ with clinical findings $C_j$, \textbf{VIVID-Med} (Fig.~\ref{fig:pipeline}) comprises three components: a trainable ViT encoder $f_\theta$ that maps $I$ to token features $X \in \mathbb{R}^{L \times d_v}$; a Structured Prediction Decomposition (SPD) projector $g_\phi$ that produces decomposed semantic tokens from $X$; and a frozen LLM $\mathcal{M}$ that maps projected tokens into structured text predictions. At test time, only the optimized backbone $f_{\theta^*}$ is deployed, discarding $g_\phi$ and $\mathcal{M}$.

\subsection{Unified Medical Schema (UMS) Supervision}
\label{sec:ums}

To avoid the semantic ambiguity of one-hot labels and free-form text, each training sample $C_j$ is converted into a Unified Medical Schema (UMS) JSON sequence encoding finding-level field-state pairs. A real example constructed from the CheXpert dataset is formatted as follows:

\begin{lstlisting}
{<@\textcolor{jsonkey}{"findings"}@>: {<@\textcolor{jsonkey}{"Lung Opacity"}@>: {<@\textcolor{jsonkey}{"state"}@>: <@\textcolor{jsonstr}{"present"}@>},
  <@\textcolor{jsonkey}{"Pneumonia"}@>: {<@\textcolor{jsonkey}{"state"}@>: <@\textcolor{jsonstr}{"uncertain"}@>},
  <@\textcolor{jsonkey}{"Pleural Effusion"}@>: {<@\textcolor{jsonkey}{"state"}@>: <@\textcolor{jsonstr}{"present"}@>},
  <@\textcolor{jsonkey}{"Cardiomegaly"}@>: {<@\textcolor{jsonkey}{"state"}@>: <@\textcolor{jsonkw}{null}@>}},
 <@\textcolor{jsonkey}{"answerability"}@>: {<@\textcolor{jsonkey}{"Lung Opacity"}@>: <@\textcolor{jsonkw}{true}@>,
  <@\textcolor{jsonkey}{"Pneumonia"}@>: <@\textcolor{jsonkw}{true}@>, <@\textcolor{jsonkey}{"Pleural Effusion"}@>: <@\textcolor{jsonkw}{true}@>,
  <@\textcolor{jsonkey}{"Cardiomegaly"}@>: <@\textcolor{jsonkw}{false}@>}}
\end{lstlisting}

\noindent Each finding's \texttt{state} $\in$ \{\texttt{present}, \texttt{absent}, \texttt{uncertain}, \texttt{null}\}, where \texttt{null} indicates the finding is not assessable from $I_j$. The \texttt{answerability} boolean mask indicates whether each finding has a valid label. This formulation drives two training mechanisms:

\textbf{Field Query Training.} Each forward pass randomly samples 4--6 finding fields per image, with elevated sampling probability (0.6) for low-frequency findings to ensure robust coverage across the long-tail distribution.

\textbf{Answerability-Aware Masking.} To prevent the network from learning noisy gradients from unassessable findings, we utilize the \texttt{answerability} field to formulate a dynamic loss weight $w_t$ for each token $t$ in the target sequence:
\[
w_t =
\begin{cases}
1, & \text{if } t \in \Omega_{\text{answerable}}, \\
0, & \text{otherwise}.
\end{cases}
\]

During training, the frozen LLM $\mathcal{M}$ receives projected visual tokens followed by a task instruction. We employ teacher forcing: the ground-truth UMS-JSON sequence is provided directly, and $f_\theta$, $g_\phi$ are optimized via next-token prediction loss, bypassing autoregressive decoding.

\subsection{Structured Prediction Decomposition (SPD)}
\label{sec:spd}

The SPD projector $g_\phi$ decomposes $X$ into $G$ complementary semantic groups. For each group $g \in \{1, \dots, G\}$, learnable queries $Q_g \in \mathbb{R}^{M \times d_v}$ perform cross-attention over $X$:
\[
A_g = \mathrm{softmax}\!\left(\frac{(Q_g W_Q)(X W_K)^\top}{\sqrt{d_h}}\right),\quad
T_g = A_g (X W_V),
\]
where $W_Q, W_K, W_V$ are learnable projections shared across all groups. We set $G=4$, $M=2$ (8 SPD tokens total). All SPD tokens and selected patch tokens are projected by a shared MLP to match the LLM embedding dimension.

To force the query groups to attend to diverse and complementary visual aspects, we impose an orthogonality regularizer on the attention maps:
\[
\mathcal{L}_{\text{ortho}} = \sum_{g \neq g'} \left\| A_g A_{g'}^\top \right\|_F^2.
\]

\begin{figure}[t]
\centering
\includegraphics[width=\textwidth]{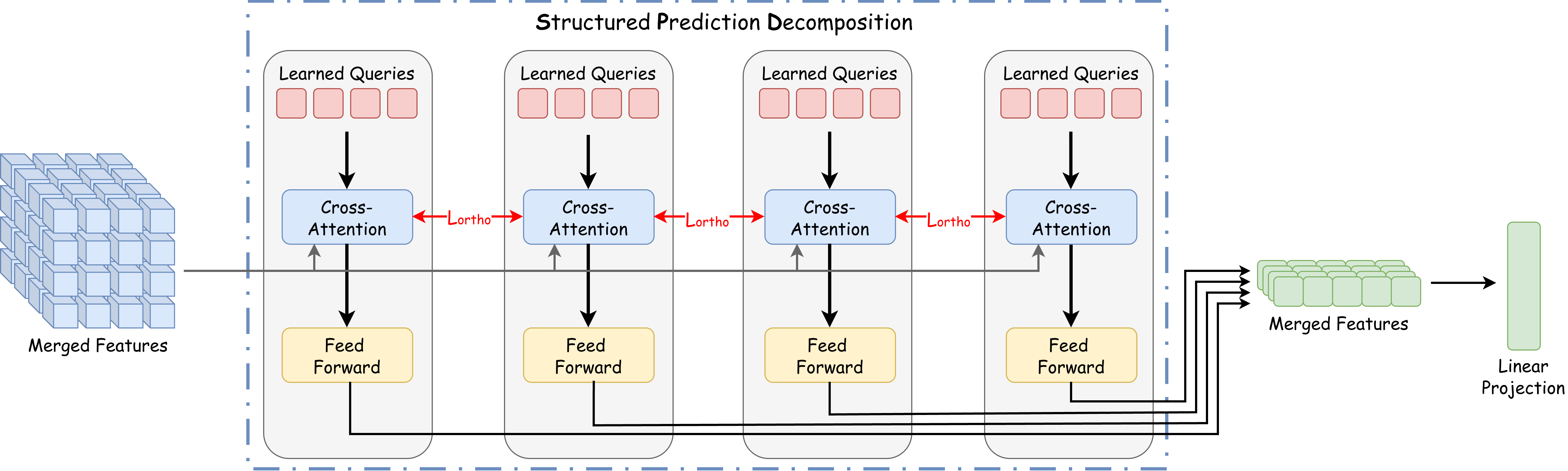}
\caption{Structured Prediction Decomposition (SPD). Multiple query groups perform cross-attention over shared ViT tokens. Orthogonality regularization encourages complementary branches before shared projection to the LLM embedding space.}
\label{fig:spd}
\end{figure}

\begin{figure}[t]
\centering
\includegraphics[width=\textwidth]{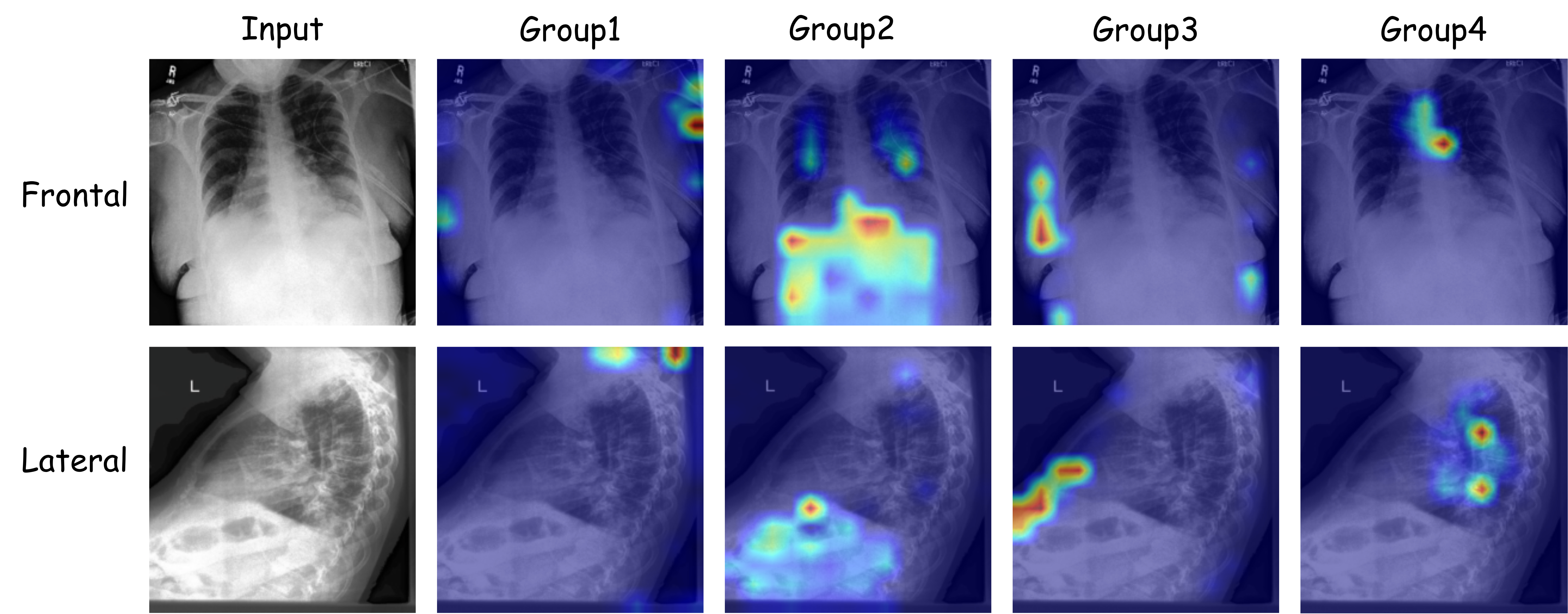}
\caption{SPD cross-attention maps for a frontal (top) and lateral (bottom) CXR. Each group attends to distinct anatomical regions.}
\label{fig:attn_maps}
\end{figure}

As shown in Fig.~\ref{fig:attn_maps}, each group consistently specializes in distinct anatomical structures across both frontal and lateral views, confirming that the orthogonality regularization drives complementary feature extraction.

\subsection{Training Objective and Inference}
\label{sec:training}

At training time, we jointly optimize the parameters of the ViT ($ \theta $) and the SPD projector ($ \phi $). The base objective is the answerability-weighted next-token prediction loss over the UMS-JSON sequence $y$:
\[
\mathcal{L}_{\text{tok}} = - \sum_t w_t \log p(y_t \mid y_{<t}, I_j).
\]
The final optimization target combines this token loss with the structural constraint:
\[
\mathcal{L} = \mathcal{L}_{\text{tok}} + \lambda_{\text{ortho}} \mathcal{L}_{\text{ortho}},
\]
where we empirically set $\lambda_{\text{ortho}}=0.01$. 

\textbf{Deployment.} After training, we discard $g_\phi$ and $\mathcal{M}$. The released artifact is a standalone ViT backbone $f_{\theta^*}$ that integrates with task-specific heads (linear probing or fine-tuning), avoiding LLM inference cost entirely.

\section{Experiments and Results}

We evaluated \textbf{VIVID-Med} across three medical image representation paradigms: in-domain classification, cross-domain transfer, and cross-modality generalization. We pretrained and evaluated on CheXpert~\cite{irvin2019chexpert} (30k CXRs, 12-label linear probing, excluding \emph{No Finding} and \emph{Pleural Other}). To assess cross-domain robustness zero-shot, we evaluated this CheXpert-trained model directly on NIH ChestX-ray14~\cite{wang2017chestx} (8 shared labels, 25,596 samples). Finally, to test cross-modality generalization without any CT pretraining, we evaluated on two CT datasets: LIDC-IDRI~\cite{armato2011lung} (slice-level benign/malignant classification) and OrganAMNIST~\cite{yang2023medmnist} (11-way organ classification).

\textbf{Implementation.} We utilized a \texttt{vit\_base\_patch16\_224} backbone ($\sim$86M parameters) and a frozen Qwen2.5-1.5B-Instruct teacher~\cite{qwen2025qwen25technicalreport}. The SPD projector (4 groups, 2 tokens/group, 4-head cross-attention, 2-layer MLP, $\sim$6M parameters) maps features to the LLM dimension ($1536$). Models were trained for up to 10k steps (effective batch size 32) using AdamW, BF16, and a cosine schedule (peak LRs: $2{\times}10^{-5}$ for ViT, $10^{-4}$ for SPD). We set $\lambda_{\text{ortho}}=0.01$. During field query training, $k \in [4,6]$ fields were sampled per image, oversampling low-frequency findings (0.6 probability) to mitigate imbalance. For evaluation, linear classifiers were trained for 3,000 steps over the frozen ViT backbones.

\textbf{Baselines.} We compare against generic visual pretraining (ImageNet-super\-vised~\cite{dosovitskiy2020image}, MAE~\cite{he2022masked}, DINOv3~\cite{simeoni2025dinov3}) and medical vision-language models (BiomedCLIP~\cite{zhang2023biomedclip}, pretrained on 15M pairs). We also include a \emph{Random-mask proxy} (simulating ViTP~\cite{li2025visual} via 75\% random token dropping instead of SPD) in the main comparison. Internal ablation variants (\emph{VIVID-Med (no SPD)}, \emph{Q-Former proxy}) are reported separately in Table~\ref{tab:ablation} to isolate each component's contribution.

\begin{table}[t]
\caption{Quantitative performance of \textbf{VIVID-Med} and baselines on in-domain (CheXpert) and cross-domain (NIH) tasks. Results are reported as mean$\pm$std over 3 seeds. NIH results are evaluated purely zero-shot without retraining.}
\label{tab:main_combined}
\centering
\small
\resizebox{\textwidth}{!}{
\begin{tabular}{lcc|cc}
\toprule
\multirow{2}{*}{Method}
& \multicolumn{2}{c|}{CheXpert-12 (in-domain)}
& \multicolumn{2}{c}{NIH-8 (cross-domain)} \\
\cmidrule(lr){2-3}\cmidrule(lr){4-5}
& Macro-AUC & Macro-F1 & Macro-AUC & Macro-F1 \\
\midrule
ImageNet supervised~\cite{dosovitskiy2020image} & $0.7746\std{0.0081}$ & $0.8813\std{0.0046}$ & $0.6147\std{0.0012}$ & $0.2090\std{0.0111}$ \\
MAE~\cite{he2022masked} & $0.7211\std{0.0031}$ & $0.8657\std{0.0096}$ & $0.5945\std{0.0013}$ & $0.1847\std{0.0005}$ \\
DINOv3~\cite{simeoni2025dinov3} & $0.6876\std{0.0053}$ & $0.8760\std{0.0014}$ & $0.6287\std{0.0023}$ & $0.2049\std{0.0050}$ \\
BiomedCLIP~\cite{zhang2023biomedclip} & $\underline{0.7923\std{0.0249}}$ & $\underline{0.8989\std{0.0050}}$ & $\underline{0.6725\std{0.0076}}$ & $\underline{0.2246\std{0.0058}}$ \\
Random-mask proxy (75\%) & $0.7396\std{0.0020}$ & $0.8705\std{0.0156}$ & $0.6000\std{0.0018}$ & $0.1969\std{0.0031}$ \\
\midrule
\rowcolor{gray!15} \textbf{VIVID-Med} (ours) & $\mathbf{0.8588\std{0.0033}}$ & $\mathbf{0.9016\std{0.0060}}$ & $\mathbf{0.7225\std{0.0019}}$ & $\mathbf{0.2621\std{0.0010}}$ \\
\bottomrule
\end{tabular}
}
\end{table}

\subsection{CXR Representation and Cross-Domain Transfer}

Table~\ref{tab:main_combined} summarizes performance on CheXpert and NIH. In-domain, \textbf{VIVID-Med} achieves an exceptional 0.8588 macro-AUC on CheXpert, outperforming BiomedCLIP by {+}6.65 points despite using 500$\times$ less pretraining data. Conversely, self-supervised baselines (MAE, DINOv3) show competitive F1s but substantially lower AUCs, highlighting their inability to capture clinical ranking structures without explicit semantic guidance.

\begin{figure}[t]
\centering
\includegraphics[width=\textwidth]{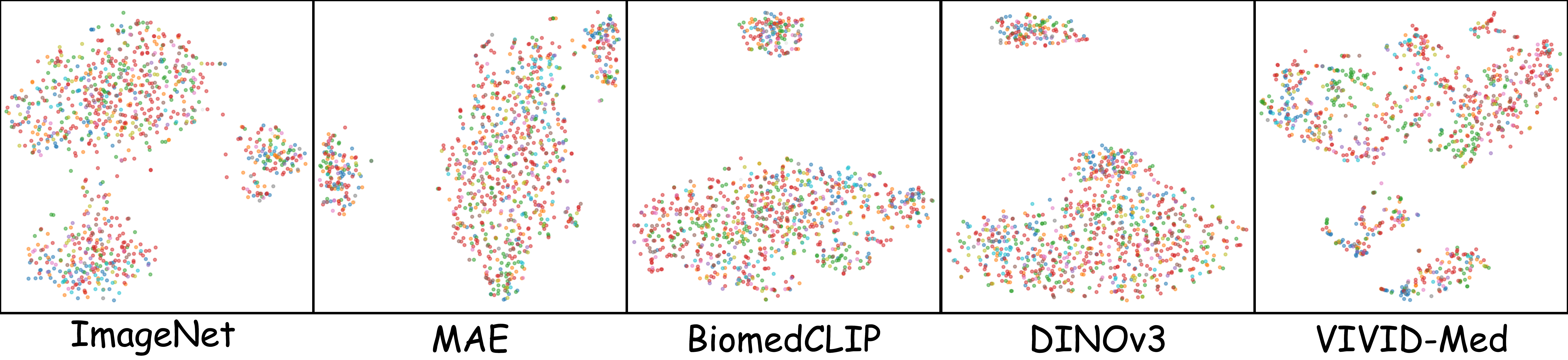}
\caption{t-SNE of CLS embeddings on CheXpert. \textbf{VIVID-Med} produces tighter, more separable clusters than ImageNet supervised and BiomedCLIP baselines.}
\label{fig:tsne}
\end{figure}

Under zero-shot transfer to NIH, \textbf{VIVID-Med} demonstrates robust out-of-distribution generalization, reaching 0.7225 macro-AUC ({+}5.00 over BiomedCLIP). Because NIH exhibits severe class imbalance ($<$5\% positives), AUC remains the more reliable metric over absolute F1. Furthermore, \textbf{VIVID-Med} significantly surpasses the Random-mask proxy, confirming that SPD's adaptive decomposition is far superior to random token dropping. Fig.~\ref{fig:tsne} further shows that the resulting CLS embeddings form tighter, more separable clusters compared to ImageNet supervised and BiomedCLIP baselines, indicating that structured LLM supervision encourages the ViT to organize its representation space according to clinically meaningful semantic distinctions.

\begin{table}[t]
\caption{Cross-modality transfer: CXR-pretrained ViT $\rightarrow$ CT slice classification (mean$\pm$std, 3 seeds). No CT data is used during pretraining.}
\label{tab:ct}
\centering
\small
\resizebox{\textwidth}{!}{
\begin{tabular}{lcc|cc}
\toprule
\multirow{2}{*}{Method}
& \multicolumn{2}{c|}{LIDC-IDRI (binary)}
& \multicolumn{2}{c}{OrganAMNIST (11-class)} \\
\cmidrule(lr){2-3}\cmidrule(lr){4-5}
& AUC & F1 & Macro-AUC & Macro-F1 \\
\midrule
ImageNet supervised~\cite{dosovitskiy2020image} & $0.7933\std{0.0336}$ & $0.6760\std{0.0251}$ & $\underline{0.9928\std{0.0001}}$ & $\underline{0.8846\std{0.0007}}$ \\
MAE~\cite{he2022masked} & $0.7607\std{0.0507}$ & $0.5341\std{0.0350}$ & $0.9826\std{0.0001}$ & $0.7743\std{0.0024}$ \\
DINOv3~\cite{simeoni2025dinov3} & $0.6675\std{0.0668}$ & $0.4389\std{0.0191}$ & $0.9221\std{0.0009}$ & $0.4570\std{0.0065}$ \\
BiomedCLIP~\cite{zhang2023biomedclip} & $\mathbf{0.8465\std{0.0148}}$ & $\underline{0.7235\std{0.0040}}$ & $0.9913\std{0.0002}$ & $0.8732\std{0.0052}$ \\
\midrule
\rowcolor{gray!15} \textbf{VIVID-Med} (ours) & $\underline{0.8413\std{0.0260}}$ & $\mathbf{0.7563\std{0.0357}}$ & $\mathbf{0.9969\std{0.0001}}$ & $\mathbf{0.9322\std{0.0004}}$ \\
\bottomrule
\end{tabular}
}
\end{table}

\subsection{Cross-Modality Transfer to CT}

Table~\ref{tab:ct} details CXR-pretrained representations evaluated on CT slices. \textbf{VIVID-Med} consistently outperforms all external baselines on OrganAMNIST despite zero CT exposure during pretraining, achieving near-perfect macro-AUC (0.9969) and macro-F1 (0.9322, {+}5.90 over BiomedCLIP), confirming that structured CXR supervision captures highly generalizable anatomical priors. DINOv3 suffers a severe drop (0.4570 F1), as its natural-image self-distillation translates poorly to medical CTs. On same-anatomy LIDC-IDRI, \textbf{VIVID-Med} achieves comparable AUC to BiomedCLIP (0.8413 vs.\ 0.8465, within one standard deviation) while substantially exceeding it in F1 ({+}3.28). The high variance on LIDC-IDRI (875 cases, std $>$0.02) reflects the inherent instability of small-scale clinical datasets, yet \textbf{VIVID-Med} still outperforms all other baselines (ImageNet, MAE, DINOv3) by a wide margin on both metrics.

\begin{table}[t]
\caption{Ablation studies isolating the impact of supervision and architecture on the CheXpert linear probe task. Results are reported as mean$\pm$std over 3 seeds.}
\label{tab:ablation}
\centering
\small
\begin{tabular}{lcccc}
\toprule
Setting & UMS & SPD & Macro-AUC & Macro-F1 \\
\midrule
\textbf{VIVID-Med} (free-text) & & & $0.8253\std{0.0028}$ & $0.8877\std{0.0027}$ \\
\textbf{VIVID-Med} (no SPD) & \checkmark & & $\underline{0.8431\std{0.0035}}$ & $0.8963\std{0.0012}$ \\
\rowcolor{gray!15} \textbf{VIVID-Med} & \checkmark & \checkmark & $\mathbf{0.8588\std{0.0033}}$ & $\underline{0.9016\std{0.0060}}$ \\
\midrule
Q-Former proxy & \checkmark & $\triangle$ & $0.8182\std{0.0195}$ & $\mathbf{0.9167\std{0.0016}}$ \\
\bottomrule
\end{tabular}
\end{table}

\subsection{Ablation Studies}

Table~\ref{tab:ablation} isolates the impact of our framework's components. Replacing free-form text supervision with structured UMS JSON improves macro-AUC by {+}1.78 points. Adding SPD adds another {+}1.57 points, netting a {+}3.35 improvement over the free-text baseline. We also evaluated a Q-Former proxy (one 8-token group, no orthogonality). While it marginally improves F1, it yields a substantially lower AUC (0.8182) with high variance. Per-class analysis confirms SPD's gains are heavily concentrated on long-tail findings (e.g., Pneumonia {+}3.9, Lung Lesion {+}3.6 AUC points), proving orthogonal decomposition fundamentally improves tail-class ranking.

\section{Discussion and Conclusion}

Results demonstrate that \textbf{VIVID-Med} achieves strong representation transfer consistently across in-domain, cross-domain, and cross-modality settings, with improvements most pronounced in macro-AUC. \textbf{VIVID-Med} offers a new perspective on medical vision-language pretraining by decoupling semantic supervision from the deployment architecture: a frozen LLM defines a stable target manifold, and the ViT explicitly learns to predict structured tokens within this space. Ablation results, particularly the Q-Former proxy's high-F1/low-AUC pattern (Table~\ref{tab:ablation}), confirm that our structured prediction decomposition (SPD) strategy is crucial for success, improving global ranking quality rather than merely fitting source-specific thresholds.

Because structured supervision aligns visual features with a continuous semantic space, \textbf{VIVID-Med} captures transferable visual priors well beyond the source modality. As demonstrated by the CT results (Table~\ref{tab:ct}), \textbf{VIVID-Med} achieves strong performance without any CT data during pretraining (0.9969 AUC on OrganAMNIST and 0.8413 AUC on LIDC-IDRI). While OrganAMNIST AUC is near saturation across most baselines ($>$0.99), the substantial F1 gap ({+}5.90 over BiomedCLIP) highlights the discriminative superiority of our method. On LIDC-IDRI, the high variance inherent to small-scale clinical datasets (875 cases) limits absolute AUC gains, yet \textbf{VIVID-Med} still outperforms all non-medical baselines by a wide margin.

\textbf{VIVID-Med} learns robust features, is resource-friendly, and effectively distills structured semantic knowledge. At deployment, only the $\sim$86M-parameter ViT backbone is retained, discarding the $\sim$1.5B-parameter LLM pipeline and thereby substantially reducing inference costs. In an age where medical AI often requires increasing resource and computational investments, \textbf{VIVID-Med} provides a fresh, scalable alternative. A clean ablation chain confirms that each component contributes independently, demonstrating that verifiable structured supervision is a highly effective route to robust medical visual representations without the need to retain massive LLMs at inference.
\bibliographystyle{splncs04}
\bibliography{ref}
\end{document}